\documentclass[sigconf]{acmart}

\AtBeginDocument{%
  }

\setcopyright{none}
\copyrightyear{2026}
\acmYear{2026}
\acmConference[DAI '26]{8th International Conference on Distributed Artificial Intelligence}{November 29--December 2, 2026}{Hong Kong}
\acmBooktitle{8th International Conference on Distributed Artificial Intelligence (DAI '26), November 29--December 2, 2026, Hong Kong}
\title{Adversarial Course-of-Action Generation: Game-Theoretic Multi-Agent Algorithms for COA matching \& COA generation}

\author{Natan Vidra}
\affiliation{%
  \institution{Anote AI}
  \city{New York}
  \country{United States}}
\email{nvidra@anote.ai}

\author{Alina Kapanova}
\affiliation{%
  \institution{Anote AI, Cornell University}
  \city{New York}
  \country{United States}}
\email{ak2765@cornell.edu}

\author{Arun Kanhai}
\affiliation{%
  \institution{Anote AI, CUNY}
  \city{New York}
  \country{United States}}
\email{arun.kanhai55@qmail.cuny.edu}

\author{Spurthi Setty}
\affiliation{%
  \institution{Stevens Institute of Technology}
  \city{New York}
  \country{United States}}
\email{ssetty2@stevens.edu}
\renewcommand{\shortauthors}{Vidra et al.}

\begin{document}

\begin{abstract}
Course-of-action (COA) generation is a distributed planning problem: a system must propose structured candidate actions, evaluate them against an adversarial response, and surface options that remain tactically coherent under changing conditions. We present COA-Bench, a small offline benchmark and reproducibility artifact for comparing COA generation policies through self-play. Following the BattleCOA terminology, we reserve COA matching for asset-effect matching and COA generation for course-of-action generation; the present artifact does not implement either DecisionFunction directly. Instead, it represents COAs as typed action chains with conditional branches, assigns a synthetic COA quality score, compares opposing COAs with a BLUE-vs-RED advantage score and Nash-gap distance, and scores doctrinal coherence with an FM 3-0-inspired heuristic rubric. Across 50 synthetic scenarios spanning five operational templates, a sampled best-response policy that draws eight RED candidates reduces BLUE advantage from .516 to .485 and BLUE wargame win rate from .920 to .820; a two-stage multi-agent council with five BLUE proposer agents, RED-team adjudication, and critique-driven revision obtains .509 BLUE advantage and .820 BLUE win rate. We also identify and fix a benchmark-design issue in which scenario framing was stored as metadata but had no effect on generated COA content. COA-Bench is not an operational battle-management system and uses no real, classified, proprietary, or human-subject data. The contribution is an inspectable evaluation harness, preliminary benchmark evidence, and lessons for building auditable agentic planning artifacts.
\end{abstract}



\maketitle

\section{Introduction}

Course-of-action development asks a planning system to propose and compare possible actions under adversarial response. In practical settings, the quality of a COA is not a property of the plan alone. It depends on the available force package, the opponent's response, timing and resource constraints, and qualitative planning criteria such as combined-arms balance, sustainment, risk mitigation, and tempo.

This paper presents COA-Bench as an industry-track benchmark report. The artifact is intentionally modest: it is offline, synthetic, deterministic where possible, and designed to expose assumptions rather than hide them behind a large model. It supports the evaluation of COA generation policies as reusable software components, records structured metrics, and makes failures inspectable.

The contribution is not an operational planner. Instead, COA-Bench makes three industry-facing contributions: (1) a formal adversarial COA-generation setup connecting asset-effect matching, course-of-action generation, and best-response evaluation without conflating those DecisionFunctions with benchmark metrics; (2) a reproducible offline harness with typed COA traces, BLUE/RED policy comparison, and doctrine-inspired diagnostics; and (3) empirical evidence that sampled adversarial response changes win-rate and plan-quality conclusions relative to single-response evaluation.

Concretely, the artifact provides:

\begin{itemize}
  \item a typed representation of synthetic COAs as action chains, conditional branches, objectives, force assignments, and optional tool calls;
  \item a self-play protocol comparing single-sample, sampled best-response, doctrine-aware response, and multi-agent council policies;
  \item a multi-metric evaluation suite including synthetic quality, BLUE-vs-RED advantage, Nash-gap distance, heuristic doctrinal alignment, candidate diversity, Pareto-frontier analysis, and a stylized wargame check;
  \item a clear scope boundary for BattleCOA-inspired benchmark metrics; and
  \item a reproducibility package that regenerates the reported results without API keys, live systems, or sensitive data.
\end{itemize}

The implementation, experiment script, and result artifacts are available at \url{https://github.com/anote-ai/research-coageneration}.

\section{BattleCOA Scope}

The BattleCOA Boot Camp material motivates a richer decision architecture in which candidate plans are represented as event structures with assets, effects, paths, and supporting relationships. COA-Bench uses that framing only as inspiration for a reproducible benchmark. In this paper, the scalar assigned to a generated COA is a \emph{synthetic COA quality score}. The pairwise score comparing BLUE and RED is a \emph{BLUE-vs-RED advantage score}. These are internal benchmark metrics, not claims about operational BattleCOA decision functions.

This distinction is important for an industry venue. COA-Bench is a research harness for reproducible evaluation, not a claim to implement the full Transformational Model for Decision Advantage or any operational battle-management workflow.

\section{Problem Formulation}

Let $s$ denote a synthetic scenario, $\pi_B$ a BLUE COA-generation policy, and $\pi_R$ a RED response policy. A policy maps a scenario and side label to a structured COA $c$. Given paired COAs, the benchmark computes a utility-like advantage signal $U(c_B,c_R,s)$, doctrine-inspired diagnostics $D(c,s)$, and candidate-set statistics such as diversity and Pareto-frontier size. The evaluation question is whether a BLUE policy that appears strong against a single response remains strong when RED samples a response set $C_R=\{c_R^1,\ldots,c_R^k\}$ and selects the highest-scoring adversarial candidate. This formalization treats COA generation as a meta-decision problem over options and responses rather than a one-shot text-generation task.

For the multi-agent setting, BLUE proposer agents produce candidates $B^0=\{b_1,\ldots,b_m\}$ and RED-team agents produce responses $R_i^0$ for each $b_i$. An adjudicator scores each candidate with:

\begin{equation}
S(b_i) = \alpha Q(b_i)+\beta D(b_i)+\gamma V(b_i,B)-\rho P(b_i,R_i),
\end{equation}

where $Q$ is rescaled synthetic quality, $D$ is doctrine proxy, $V$ is candidate-set diversity, and $P$ is adversarial pressure from the strongest RED response. The top $h=2$ candidates are revised by adding mitigation actions keyed to the strongest RED critique, producing $B^1$. The final selection is:

\begin{equation}
b^\star = \arg\max_{b \in B^0 \cup B^1} S(b).
\end{equation}

We report a robustness margin $M=S(b^\star)-\max_{b\in B^0}S(b)$ and pressure reduction $\Delta P=\bar{P}_{\mathrm{shortlist}}-P(b^\star,R^\star)$. The reported council uses five proposer roles, RED response budget four per candidate, revision width two, and weights $(.45,.25,.10,.20)$.

\section{Related Work}

\textbf{Self-play and adversarial evaluation.} AlphaZero~\cite{silver2018alphazero}, AlphaStar~\cite{vinyals2019alphastar}, and Pluribus~\cite{brown2019superhuman} show that self-play can discover strong policies in adversarial games. CICERO extends strategic reasoning into a natural-language negotiation setting~\cite{bakhtin2022cicero}. COA-Bench borrows the paired-response structure but does not train a policy; it evaluates simple generation policies in a small synthetic action space.

\textbf{Military decision modeling.} Classic attrition equations~\cite{lanchester1916} and Schelling's account of strategic conflict~\cite{schelling1960} are foundational abstractions for adversarial military interaction. COA-Bench uses a stylized attrition check only as a secondary diagnostic; it is not a validated combat model. The doctrinal-alignment rubric draws inspiration from FM 3-0 planning criteria~\cite{fm30}, but remains a heuristic feature extractor rather than a doctrine-expert judgment.

\textbf{Agent and planning benchmarks.} ReAct~\cite{yao2023react} and AgentBench~\cite{liu2024agentbench} motivate evaluating agents as systems that interleave planning, action, and feedback. SWE-bench grounds agent evaluation in executable software tasks~\cite{jimenez2024swebench}, while $\tau$-bench and ToolSandbox evaluate stateful tool-using agents in domain or conversational settings~\cite{yao2024taubench,lu2025toolsandbox}. PlanningArena and ACPBench further emphasize that planning systems should be evaluated across action selection, tool use, and state-change reasoning rather than only final text quality~\cite{zheng2025planningarena,kokel2025acpbench}. AgentDiagnose argues that trajectory-level traces expose decomposition, observation, verification, and backtracking failures hidden by aggregate success~\cite{ou2025agentdiagnose}. COA-Bench adopts the same evaluation philosophy for COA generation: it records candidate routes, quality, doctrinal features, diversity, and wargame outcomes so that policy behavior remains inspectable.

\textbf{LLM planning and tool orchestration.} Toolformer and ToolLLM study how language models learn to invoke external tools~\cite{schick2023toolformer,qin2024toolllm}. LLMCompiler optimizes function-call plans for latency and dependency structure~\cite{kim2024llmcompiler}, and SPIRAL combines symbolic planning with grounded reflective search~\cite{zhang2026spiral}. These projects motivate representing COAs as structured action artifacts rather than untyped prose. COA-Bench is not a tool-learning benchmark, but its typed actions, conditional branches, and optional tool-call slots are designed so later LLM-backed policies can be compared against simple programmatic baselines under the same scoring interface.

\section{Benchmark Design}

\subsection{COA Representation}

A COA is represented as a typed object with an objective, force assignment, domain tag, ordered or chained actions, optional conditional branch, and optional tool calls. Actions include a category, target, priority, and supporting metadata. Conditional branches allow the benchmark to represent plan structures such as reconnaissance gating a strike-or-withdraw decision.

Each COA receives a synthetic quality score in $[-1,1]$ combining effectiveness, cost, and risk:

\begin{equation}
q = w_e e - w_c c - w_r r,
\end{equation}

with default weights $w_e=.5$, $w_c=.3$, and $w_r=.2$. This score is an internal benchmark metric, not asset-effect matching.

\subsection{Pairwise Evaluation}

For paired BLUE and RED COAs, we report a BLUE-vs-RED advantage score:

\begin{equation}
\mathrm{Advantage} = (q_{\mathrm{blue}} - q_{\mathrm{red}} + 2) / 4.
\end{equation}

We also report Nash-gap distance after rescaling quality scores to $[0,1]$ payoffs. The gap is a diagnostic for imbalance in the paired exchange, not evidence of convergence to equilibrium.

\begin{table}[t]
\caption{Adversarial COA evaluation protocol.}
\label{tab:algorithm}
\small
\begin{tabular}{@{}p{.96\linewidth}@{}}
\toprule
\textbf{Input:} scenario $s$, BLUE policy $\pi_B$, RED policy $\pi_R$, response budget $k$ \\
1. Generate BLUE candidate $c_B \leftarrow \pi_B(s)$. \\
2. Sample RED candidates $C_R=\{c_R^1,\ldots,c_R^k\}$ from $\pi_R(s,c_B)$. \\
3. Score each pair with advantage $U(c_B,c_R^i,s)$ and diagnostics $D(c_R^i,s)$. \\
4. Select RED best response and record diversity, Pareto frontier, and trace metrics. \\
5. Return ranked COA pair, adversarial score, and inspectable evaluation trace. \\
\bottomrule
\end{tabular}
\end{table}

\subsection{Doctrinal and Wargame Diagnostics}

The heuristic doctrinal rubric scores objective clarity, intelligence preparation, combined-arms balance, sustainment, risk mitigation, and tempo or sequencing. A separate stylized Lanchester-style wargame check scales raw combat power by synthetic quality and doctrinal alignment. Both diagnostics are intentionally transparent and limited.

\section{Evaluation}

\subsection{Protocol}

We generate 50 scenarios from 10 base seeds and five templates: urban stability operations, maritime interdiction, multi-domain combat, suppression of enemy air defenses, and humanitarian evacuation. The two added templates stress different parts of the artifact. The air-defense case requires coordinated ISR, electromagnetic/cyber activity, and time-sensitive kinetic sequencing. The humanitarian evacuation case requires security, logistics, medical staging, and public-information actions under higher law-of-armed-conflict ambiguity. For each scenario, we keep the first BLUE seed COA fixed and generate RED responses in three ways:

\begin{enumerate}
  \item a single random best response from \texttt{SelfPlayEngine}; and
  \item a sampled best response that draws $n=8$ candidates and returns the highest-quality one; and
  \item a doctrine-aware sampled response that selects by a weighted quality/doctrine proxy.
  \item a two-stage multi-agent council in which five BLUE proposer agents generate role-specialized COAs, RED-team agents stress-test each candidate, and an adjudicator revises the top two candidates before final selection.
\end{enumerate}

We compute BLUE advantage, Nash gap, doctrinal alignment, candidate diversity, quality-score spread, Pareto-optimal candidate count, selection regret, response-budget curves, and wargame outcome. Confidence intervals are 95\% percentile bootstrap intervals over the 50 scenarios.

\begin{table}[t]
\caption{RED response policies over 50 scenarios.}
\label{tab:main}
\small
\begin{tabular}{lrrr}
\toprule
Metric & Single & Quality & Doctrine \\
\midrule
BLUE advantage & .516 & .485 & .488 \\
Nash gap & .194 & .255 & .250 \\
RED doctrinal alignment & .722 & .731 & .753 \\
BLUE win rate & .920 & .820 & .800 \\
\bottomrule
\end{tabular}
\end{table}

\begin{table}[t]
\caption{Response-budget curve for quality-greedy RED.}
\label{tab:budget}
\small
\begin{tabular}{lrrrrr}
\toprule
Budget $k$ & 1 & 2 & 4 & 8 & 16 \\
\midrule
BLUE advantage & .516 & .498 & .490 & .485 & .479 \\
\bottomrule
\end{tabular}
\end{table}

\begin{table}[t]
\caption{Multi-agent council over 50 scenarios.}
\label{tab:council}
\small
\begin{tabular}{lr}
\toprule
Metric & Value \\
\midrule
BLUE advantage & .509 [.504, .514] \\
BLUE win rate & .820 \\
Council diversity & .571 [.493, .650] \\
Revised diversity & .644 [.593, .701] \\
Consensus gap & .012 [.009, .016] \\
Adversarial pressure & .615 [.607, .624] \\
Robustness margin & .014 [.010, .018] \\
Pressure reduction & .002 [-.005, .010] \\
Revised selected rate & .760 \\
\bottomrule
\end{tabular}
\end{table}

\subsection{Results}

Table~\ref{tab:main} shows that sampled best-response makes RED a tougher opponent. BLUE advantage falls from .516 to .485 because RED's quality score increases under best-of-eight selection. The stylized wargame check points in the same direction: BLUE win rate falls from .920 to .820. A doctrine-aware selector produces a similar adversarial effect (.488 advantage, .800 BLUE win rate) while making the quality/doctrine tradeoff explicit.

The sampled candidate set is also useful before selection. Mean candidate diversity is .682, mean quality-score spread is .246, and the mean number of Pareto-optimal candidates is 2.16 out of 8. The doctrine-aware selector changes the chosen response in 56\% of scenarios; relative to the quality-greedy response, this incurs .041 quality regret while gaining .106 doctrinal alignment. Table~\ref{tab:budget} shows increasing adversarial pressure as RED samples more candidates: BLUE advantage drops from .516 at $k=1$ to .498 at $k=2$, .490 at $k=4$, .485 at $k=8$, and .479 at $k=16$.

Table~\ref{tab:council} reports the new two-stage multi-agent council experiment. Unlike the RED-only response policies, the council changes the BLUE generation process itself: maneuver, intelligence, cyber, sustainment, and protection proposer agents each produce a BLUE candidate, RED-team agents respond to each one, the adjudicator shortlists two candidates, and a revision operator adds mitigation actions keyed to the strongest RED critique. The council raises BLUE advantage relative to quality-greedy adversarial evaluation (.509 vs. .485) and produces a BLUE win rate of .820. Revised candidates are selected in 76\% of scenarios, showing that the second-stage deliberation is not cosmetic. Revised-candidate diversity (.644) exceeds initial council diversity (.571), but pressure reduction is small and its interval overlaps zero; the main measured benefit is therefore candidate repair and selection robustness rather than clear avoidance of RED pressure.

\subsection{Implementation Artifact}

The implementation separates three layers. The scenario layer constructs typed synthetic cases and seed BLUE COAs. The policy layer exposes interchangeable response policies: a single-sample baseline, a quality-greedy sampled response, and a doctrine-aware sampled response. The evaluation layer consumes generated BLUE/RED pairs and writes traceable result artifacts. This separation is useful for an industry benchmark because future teams can replace a policy while preserving the same scenario corpus, metrics, and reporting format.

\begin{table}[t]
\caption{Generated reproducibility artifacts.}
\label{tab:artifacts}
\small
\begin{tabular}{lp{.58\linewidth}}
\toprule
Artifact & Purpose \\
\midrule
\texttt{rows.csv} & Per-scenario policy outcomes and diagnostics. \\
\texttt{summary.csv} & Bootstrap means and confidence intervals. \\
\texttt{budget\_curve.csv} & Sensitivity to RED response budget $k$. \\
\texttt{council\_rows.csv} & Multi-agent proposer/RED-team/adjudicator traces. \\
\texttt{terrain\_summary.csv} & Per-template averages for corpus inspection. \\
\texttt{details.json} & Run metadata, win rates, and configuration. \\
\bottomrule
\end{tabular}
\end{table}

The artifact is deliberately file-based rather than service-based. This avoids hidden state, API dependence, and environment-specific model behavior. It also lets reviewers inspect whether a headline result is driven by all scenarios or by one scenario family. For example, the terrain summary now separates urban, maritime, mixed-domain, air-defense, and humanitarian-evacuation cases. That breakdown is not presented as operational evidence; it is a guardrail against accidentally overfitting a benchmark claim to one synthetic template.

Finally, the response-budget curve is included as a first-class output rather than a one-off plot. In a deployed planning workflow, an adversary with more time or compute can search more responses. Treating $k$ as an explicit experimental variable makes that assumption visible and lets the benchmark distinguish a policy that is robust to stronger response search from one that only performs well against shallow opposition.

\subsection{Benchmark-Design Finding}

During development, we found that the scenario-framing parameter was a silent no-op. It was stored as metadata but did not change objective text or action content. A naive framing-sensitivity test therefore returned exactly zero change. We patched the generator so framing affects the objective and, under a BLUE-favorable framing, can add an action category that changes the combined-arms-balance term. Across the expanded corpus, the resulting framing-sensitivity delta is .066 [.049, .086]. We report this directly as a benchmark-design finding rather than treating it as a realistic estimate of framing effects.

\section{Industry Lessons}

First, benchmark terminology must be controlled. Reusing COA matching and COA generation for internal metrics created a misleading connection to BattleCOA DecisionFunctions. The correction now distinguishes formal decision functions from synthetic evaluation scores.

Second, adversarial sampling changes conclusions. A COA that looks strong against one response may look weaker when the opponent samples several plausible responses.

Third, multi-COA output is more useful than one selected plan when candidates include diverse, nonredundant options.

Fourth, reproducibility artifacts help catch evaluation bugs. The framing no-op was visible because the pipeline exposed scenario metadata, generated content, and metric outputs separately.

Fifth, industry-facing agentic planning benchmarks need explicit observability. The practical value of COA-Bench is less the absolute score of any current policy and more the ability to inspect why a policy looked strong: which candidates were generated, how much diversity they contained, which doctrinal rubric terms saturated, and whether a scenario variable actually affected generated content. This aligns with trajectory-diagnosis work in LLM agents~\cite{ou2025agentdiagnose} and with DAI's emphasis on artifacts, observability, and reproducible evaluation for deployed distributed AI systems.

\section{Threats to Validity and Ethics}

COA-Bench is synthetic and offline. It uses no real planner, real intelligence data, classified data, proprietary data, or human-subject data. Its main internal-validity risk is metric misspecification: the synthetic quality score, doctrinal rubric, and wargame check may reward artifacts that would not survive expert review. Its external-validity risk is scenario simplicity: 50 templated cases cannot represent the complexity of operational planning. Its construct-validity risk is that COA matching and COA generation are scoped as motivating DecisionFunctions, while the present benchmark evaluates typed COA artifacts rather than a full BattleCOA pipeline. The current evaluated policies are simple random-sampling policies; the repository contains an LLM-backed policy, but it is not evaluated in this paper.

Because COA generation concerns high-impact defense contexts, any future live evaluation would require strict data governance, human review, security controls, expert validation, and clear limits on autonomous action. The current artifact performs no external action and is intended only for research on evaluation methodology.

\section{Conclusion}

COA-Bench provides a small, inspectable benchmark for evaluating course-of-action generation policies under adversarial self-play. It shows that sampled best-response produces a tougher synthetic opponent, that candidate sets contain meaningful diversity, and that transparent artifacts can expose benchmark-design failures such as no-op scenario framing. The result is a preliminary but useful foundation for stronger BattleCOA-oriented benchmarks.

\begin{acks}
No human participants, proprietary data, classified data, or operational systems were used in the reported offline benchmark. Funding sources, organizational support, and individual acknowledgments should be completed by the authors before final submission if applicable.
\end{acks}

\section*{Generative AI Disclosure}
Generative AI tools materially assisted with software implementation, experiment scripting, terminology review, and manuscript drafting. The authors are responsible for reviewing the code, verifying the generated results against the released artifacts, checking all citations, and approving the final claims and text.

\bibliographystyle{ACM-Reference-Format}
\bibliography{references}

\end{document}